\pdfoutput=1

\documentclass[11pt]{article}

\usepackage[final]{coling}

\usepackage{times}
\usepackage{latexsym}

\usepackage[T1]{fontenc}

\usepackage[utf8]{inputenc}

\usepackage{microtype}

\usepackage{inconsolata}

\usepackage{graphicx}

\usepackage{xltabular}
\usepackage{booktabs}
\usepackage{enumitem}

\title{Financial Named Entity Recognition: How Far Can LLM Go?}

\author{Yi-Te Lu$^{1}$ \and Yintong Huo$^{2}$ \\
\\
$^{1}$ National Taiwan University, Taiwan \\
$^{2}$ Singapore Management University, Singapore \\
\\
\footnotesize{ \textbf{Correspondence:} \href{mailto:b08901016@ntu.edu.tw}{b08901016@ntu.edu.tw}}
}

\begin{document}
\maketitle
\begin{abstract}

The surge of large language models (LLMs) has revolutionized the extraction and analysis of crucial information from a growing volume of financial statements, announcements, and business news. Recognition for named entities to construct structured data poses a significant challenge in analyzing financial documents and is a foundational task for intelligent financial analytics. However, how effective are these generic LLMs and their performance under various prompts are yet need a better understanding. To fill in the blank, we present a systematic evaluation of state-of-the-art LLMs and prompting methods in the financial Named Entity Recognition (NER) problem. Specifically, our experimental results highlight their strengths and limitations, identify five representative failure types, and provide insights into their potential and challenges for domain-specific tasks.

\end{abstract}

\section{Introduction}

As an increasing amount of information is contained within documents and text available online, utilizing a series of natural language processing (NLP) techniques to automate the process of extracting meaningful information from unstructured text has become a critical task, especially in the financial domain \cite{ashtiani2023news}. Among all, named entity recognition (NER) serves as a foundational first step in identifying key entities, such as persons, organizations, and locations, enabling the construction of knowledge graphs and other applications. 

With the surge of large language models (LLMs), LLMs have demonstrated transformative capabilities in generative tasks, leveraging reinforcement learning from human feedback (RLHF) \cite{christiano2017deep}. LLMs achieve remarkable performance across a wide range of NLP tasks with minimal adaptation \cite{qin2024large}. However, their ability to perform domain-specific tasks, such as NER in the financial domain, remains less explored. For instance, in the sentence \textit{``Johnson Brothers rethink plan for St. Paul waterfront Shepard Road Development.''}, a generic NER model might incorrectly classify the company \textit{"Johnson Brothers"}as a person. This understanding is critical, as it could influence numerous applications in finance.

In this paper, we aim to evaluate the capabilities of state-of-the-art LLMs in performing NER tasks within the financial domain, their response to various prompt types, and their limitations in this context. To achieve this, we conduct a systematic analysis and present experimental results, comparing the effectiveness of leading LLMs with recent fine-tuned approaches. Specifically, we evaluate three advanced LLMs with different parameter sizes, GPT-4o \cite{openai2024chatgpt}, LLaMA-3.1 \cite{dubey2024llama}, and Gemini-1.5 \cite{gemini2024api}—under three distinct prompting techniques: direct prompting, in-context learning, and chain-of-thought (CoT) prompting. We perform our study by investigating the following two research questions (RQs):

\begin{itemize}[leftmargin=*]
    \item \textbf{RQ1}: How do different LLMs perform in NER tasks under various prompts?
    \item \textbf{RQ2}: What types of mistakes do LLMs commonly make?
\end{itemize}

To sum up, the main contributions of this paper are as follows:
\begin{itemize}[leftmargin=*]
    \item To the best of our knowledge, this is the first study to comprehensively compare state-of-the-art generically trained LLMs on NER tasks in the financial domain.
    \item We analyze LLM performance across three distinct prompting techniques, identify their limitations, categorize five representative types of failures and underlying causes, and elicit two future directions based on our findings.
\end{itemize}

\section{Related Work}

\subsection{Large Language Models in Finance}

LLMs have recently been applied to finance, particularly in automatic information retrieval and financial analysis \cite{li2023large}. \citealp{li2023chatgpt} empirically explore ChatGPT and GPT-4's capabilities in analyzing financial texts and compare them to state-of-the-art fine-tuned models. However, existing research mainly focuses on fine-tuned finance LLMs or individual generic LLMs, lacking comparisons of their performance under various prompt designs. This paper addresses this gap by providing a comprehensive evaluation of state-of-the-art LLMs under various prompting styles in the context of financial NER tasks.

\section{Study Setup}
To understand current LLMs' capabilities in handling financial NER problems, we choose three state-of-the-art LLMs, each with three popular prompting strategies. We further select two representative transformer-based models and fine-tune them on financial data for comparison.

\subsection{Financial NER Datasets}

In this study, we use the FiNER-ORD dataset \cite{shah2023finer} as our benchmark. While the CRA NER dataset \cite{alvarado2015domain}, based on financial agreements from the SEC, is widely used for research \cite{li2023chatgpt} and includes four entity types (person/PER, location/LOC, organization/ORG, and miscellaneous/MISC), it suffers from a skewed distribution of entity types and limited source of data. 

FiNER-ORD resolves this imbalance and removes the ambiguous miscellaneous category, consisting of a manually annotated dataset of 201 financial news articles. This provides a more robust and high-quality benchmark for financial NER tasks and has been adopted in recent research \cite{xie2024pixiu}. As reported by \citealp{shah2023finer}, the entity ratio in FiNER-ORD for ORG, LOC, and PER  is 2.29:1.17:1, compared to the heavily skewed ratio of 0.31:0.22:1 in the CRA dataset.

\subsection{Models}

We evaluate three state-of-the-art LLMs and their lightweight versions on the FiNER-ORD task: GPT-4o, GPT-4o-mini \cite{openai2024chatgpt}, LLaMA-3.1-70B-Instruct, LLaMA-3.1-8B-Instruct, Gemini-1.5-flash, and Gemini-1.5-flash-8B \cite{gemini2024api}. The model versions are 20240806 for GPT-4o, 20240718 for GPT-4o-mini, 20240723 for LLaMA-3.1, and the latest stable release for Gemini-1.5-flash models as of November. LLaMA-3.1 models are accessed through the DeepInfra API \cite{deepinfra}. All models use default configurations as per their respective API documentation \cite{openaichatgptguide, gemini2024api, deepinfra}.

Additionally, we evaluate transformer-based models for comparison: BERT \cite{devlin2018bert} and RoBERTa \cite{liu2019roberta}. These models are initialized with pre-trained versions available in the Hugging Face Transformers library \cite{wolf2020transformers}, using a batch size of 16, a learning rate of 1e-05, and 50 epochs. Fine-tuning is performed on an Nvidia Tesla A100 GPU via Google Colab \cite{googlecolab}.

\subsection{Prompt Design}

\begin{figure}[t]
\centering
\includegraphics[width=\linewidth]{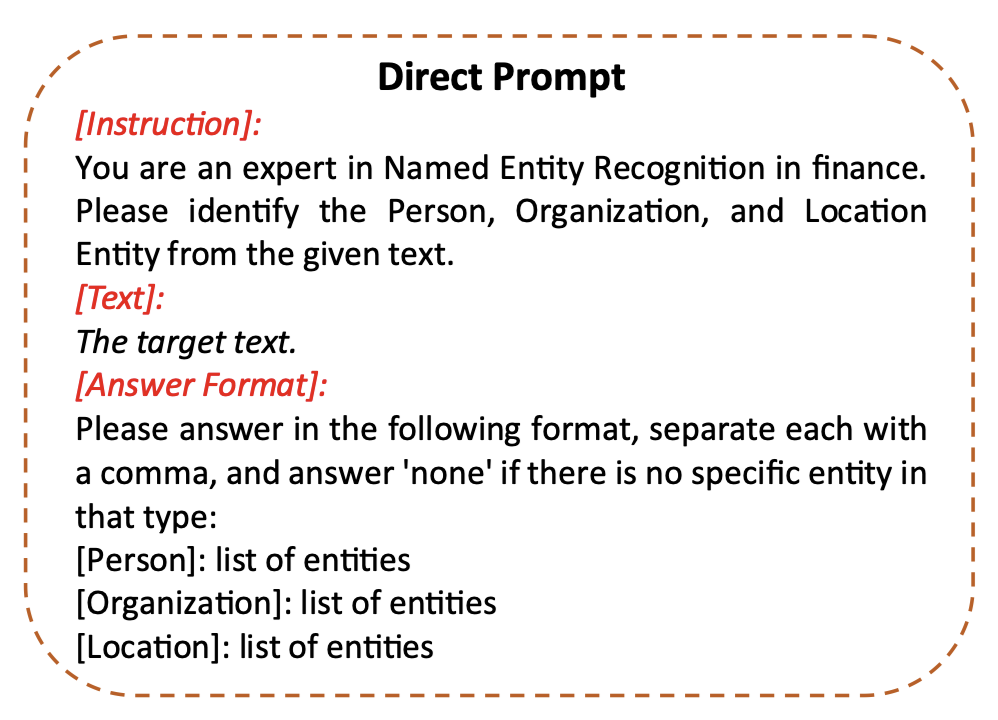}
\caption{Direct prompt for the NER task.}  
\label{fig:directprompt}
\end{figure}

\begin{figure}[t]
\centering
\includegraphics[width=0.9\linewidth]{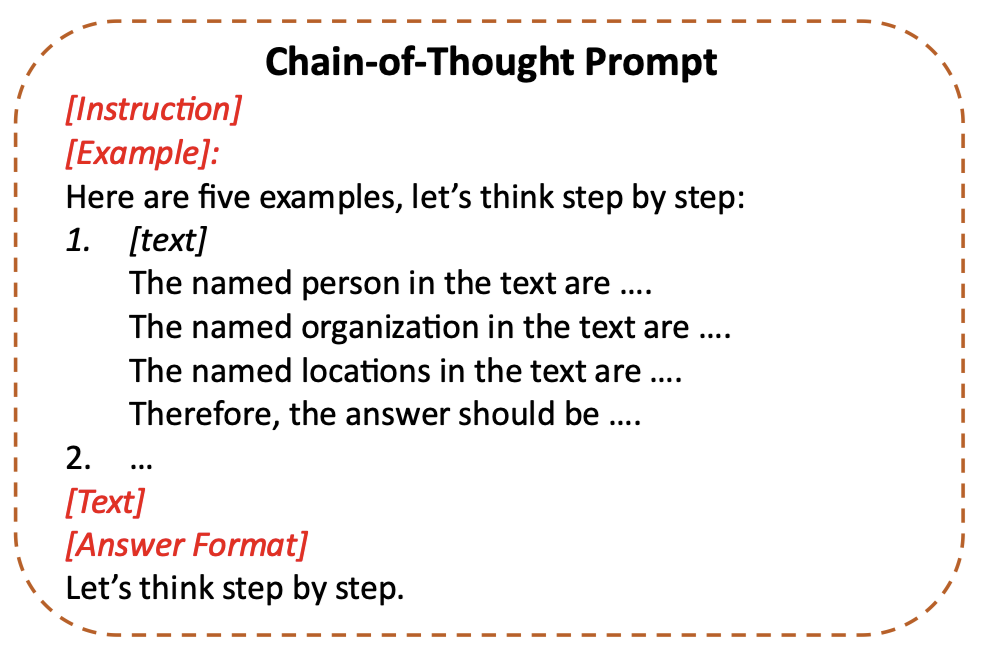}
\caption{The chain-of-thought prompt for experiments.}  
\label{fig:CoTprompt}
\end{figure}

We design three types of prompt methods: direct prompt, in-context learning \cite{dong2022survey}, and chain-of-thought \cite{wei2022chain}. As shown in Figure~\ref{fig:directprompt}, the direct prompt first gives instructions for the NER task, followed by the given text and the answer format. Next, we conduct few-shot learning (five shots) experiments through in-context learning and CoT prompts. The shots are chosen randomly and the same five shots are used in every experiment. For the in-context learning prompt, we simply add the five examples after the NER task instruction of the direct prompt. For the chain-of-thought prompt, we use the instruction "let's think step by step" to design intermediate steps for identifying each named entity in the text, as shown in Figure~\ref{fig:CoTprompt}.

\subsection{Evaluation Metrics}

After obtaining answers from the generated text, we label the identified entities through word matching. The evaluation metrics include the \textit{entity-level F1 score} and the \textit{weighted F1 score}. The formula for \textit{entity-level F1 score} is described below, where $TP$, $FP$, and $FN$ represent the counts of True Positives, False Positives, and False Negatives, respectively.
\begin{equation}
\small
    Precision = \frac{TP}{(TP + FP)}, \ Recall = \frac{TP}{(TP + FN)}
\end{equation}

\begin{equation}
\small
    F1\_Score = \frac{2*Precision*Recall}{Precision+Recall}
\end{equation}
The \textit{weighted F1 score} is defined as follows:
\begin{equation}
\small
    w_i = \frac{No.\_of\_entities\_in\_class_i}{Total\_number\_of\_entities}
\end{equation}
\begin{equation}
\small
    Weighted\_F1 = \sum\limits_{i=1}^N (w_i*F1\_Score_i)
\end{equation}

\section{Experiments}

\begin{table}[h]
\small
    \caption{Performance of different fine-tuned language models and LLMs under different prompts on FiNER-ORD task.}
    \vspace{-0.1in}
    \centering
    \label{tab:LLMresult}
    \begin{tabular}{l||cccc}
        \toprule
            Model & PER & LOC  & ORG  & Weighted  \\
        \midrule
        \multicolumn{5}{c}{Fine-Tuned Language Models} \\
        \midrule
            BERT & \textbf{0.9664} & 0.8674 & 0.8313 & 0.8744 \\
            RoBERTa & 0.9663 & \textbf{0.8748} & \textbf{0.8379} & \textbf{0.8792} \\
        \midrule
        \multicolumn{5}{c}{LLMs} \\
        \midrule
             GPT-mini & 0.8296 & 0.7669 & 0.6824 & 0.7396 \\
             LLaMA-8B & \textbf{0.8799} & \textbf{0.7973} & \textbf{0.7299} &  \textbf{0.7839} \\
             Gemini-8B & 0.8536 & 0.7773 & 0.6732 & 0.7434 \\
        \midrule
             GPT & 0.9023 & 0.8009 & \textbf{0.7312} & \textbf{0.7910} \\
             LLaMA-70B & \textbf{0.9042} & 0.7958 & 0.7073 & 0.7781 \\
             Gemini & 0.8802 & \textbf{0.8228} & 0.7238 & 0.7868 \\
        \midrule
        \multicolumn{5}{c}{Few-Shot Learning (5-shot)} \\
        \multicolumn{5}{c}{In-Context Learning} \\
        \midrule
            GPT-mini & 0.9265 & \textbf{0.8061} & 0.6841 & 0.7743 \\
            LLaMA-8B & 0.8681 & 0.7681 & 0.7132 & 0.7655 \\
            Gemini-8B & \textbf{0.9308} & 0.7991 & \textbf{0.7468} & \textbf{0.8059} \\
        \midrule
            GPT & 0.9372 & \textbf{0.8381} & 0.7541 & 0.8203 \\
            LLaMA-70B & 0.9415 & 0.7947 & 0.7948 & 0.8321 \\
            Gemini & \textbf{0.9418} & 0.8106 & \textbf{0.7966} & \textbf{0.8368} \\
        \midrule
        \multicolumn{5}{c}{Chain-of-Thought (CoT)} \\
        \midrule
            GPT-mini & 0.9221 & \textbf{0.8072} & 0.7389 & 0.8015 \\
            LLaMA-8B & 0.8467 & 0.7505 & 0.7005 & 0.7494 \\
            Gemini-8B & \textbf{0.9343} & 0.7900 & \textbf{0.7408} & \textbf{0.8016} \\
        \midrule
            GPT & 0.9361 & \textbf{0.8295} & 0.7466 & 0.8142\\
            LLaMA-70B & 0.9122 & 0.7996 & 0.7514 & 0.8036  \\
            Gemini & \textbf{0.9378} & 0.8171 & \textbf{0.7958} & \textbf{0.8369} \\
        \bottomrule
    \end{tabular}
\end{table}

In this work, we conduct experiments to answer the following two research questions.
\begin{table*}[h]
\small
    \caption{Failure types, distributions, and examples. \textcolor{blue}{Entities} and their \textcolor{red}{wrong recognitions} are highlighted with \textcolor{blue}{blue} and \textcolor{red}{red}, respectively.}
    \vspace{-0.1in}
    \centering
    \label{tab:failurecase}
    \begin{tabular}{l|c|l}
        \toprule
            Failure Type & Ratio & Example text and mislabeled entities \\
        \midrule
        Contextual & 31.3\% & \textit{\textcolor{blue}{Johnson Brothers} rethink plan for St. Paul waterfront Shepard Road Development.}\\
        misunderstanding & & The company \textit{"Johnson Brothers"} is mislabeled as a \textcolor{red}{person}. \\
        \midrule
        Pronouns and & 26.3\% & \textit{Nokia was holding exclusive talks with the \textcolor{blue}{German car makers}.} \\
        generic terms & & Non-entity \textit{"German car makers"} is mislabeled as an \textcolor{red}{organization} entity. \\
        \midrule
        Citizenship & 10.3\% & \textit{One suffered by a reported 66\% of the \textcolor{blue}{British} population.} \\
        & & Non-entity \textit{"British"} is mislabeled as a \textcolor{red}{location} entity as it relates to the UK. \\
        \midrule
        Implied entities & 10.7\% & \textit{People use \textcolor{blue}{Google Maps} or another navigation service to get to their destination .}\\
        & & Non-entity \textit{"Google Maps"} is mislabeled as an \textcolor{red}{organization} as it refers to Google.\\
        \midrule
        Entity omission & 21.4\% & \textit{Will General Motors ( NYSE : GM ) be next ?}\\
        & & \textcolor{red}{Abbreviation} entity \textit{"NYSE"} is not recognized.\\
        Boundary errors & &  \textit{Johnson Brothers rethink plan for \textcolor{blue}{St. Paul waterfront Shepard Road} Development.} \\
        & & Only \textit{"St. Paul"} is labeled instead of  \textcolor{red}{complete location}, \textit{"St. Paul waterfront Shepard Road"} \\
        \bottomrule
    \end{tabular}
\vspace{-0.1in}    
\end{table*}

\subsection{RQ1: How do different LLMs perform in FiNER-ORD tasks under different prompts?}

We present the performance results of three leading LLMs under three distinct prompts in Table~\ref{tab:LLMresult}. The results are measured using the F1 scores for three entity types and the weighted F1 score (shown in the \textit{Weighted} column). The LLMs are grouped into two sections based on their size, with \textbf{bold values} highlighting the best performance. From these results, we can draw the following observations.

\noindent \textbf{(1) Fine-tuned language models consistently outperform generic LLMs, the performance gap can be narrowed through prompt design, few-shot learning, and model size.} Table~\ref{tab:LLMresult} demonstrates that fine-tuned language models surpass generic LLMs in zero-shot direct prompting. However, the performance of generic LLMs improves significantly with diverse zero-shot prompting styles, surpassing the prompt designs proposed by \citealp{shah2023finer}. Additionally, few-shot learning and larger LLMs demonstrate notable advantages over their smaller counterparts.

\noindent \textbf{(2) Chain-of-Thought prompting has limited effect on LLMs performance and can sometimes reduce effectiveness.} While few-shot learning generally enhances generic LLMs' performance, Table~\ref{tab:LLMresult} shows that the difference between prompting styles is marginal. CoT prompting only improves the performance of the GPT-4o-mini model, whereas it significantly degrades the performance of the LLaMA 3.1 series. Notably, LLaMA 3.1 frequently suffers from "implied entities" errors, where it tends to overanalyze and tag words that merely imply a named entity. This failure type is further discussed in subsequent sections.

\noindent \textbf{(3) The Gemini series outperforms the GPT-4o and LLaMA 3.1 series in the FiNER-ORD task after few-shot learning.} The Gemini series outperforms the GPT-4o and LLaMA 3.1 series in the FiNER-ORD task after few-shot learning. Experimental results indicate a consistent performance ranking, with the Gemini series achieving the optimal performance, followed closely by the GPT-4o series. The LLaMA 3.1 series exhibits the lowest performance among the three.

\subsection{RQ2: What types of mistakes do LLMs commonly make?}

We manually annotate the failure types, summarize the limitations of LLMs, and analyze the underlying causes based on their responses, as shown in Table~\ref{tab:failurecase}. The most common failure cases include: 

\noindent \textbf{(1) Contextual misunderstanding of proper noun.} LLMs often fail to classify entities that rely on context correctly, such as domain-specific terms or ambiguous entities. For example, person names that overlap with location names, and organizational entities containing person or location names may be incorrectly categorized.

\noindent \textbf{(2) Pronouns and generic terms}. Terms such as pronouns (\textit{"he"} or \textit{"a woman"}), and generic phrases (\textit{"universities"} or \textit{"automakers"}) are sometimes misclassified as specific entities.

\noindent \textbf{(3) Citizenship Terms.} Words related to citizenship, such as \textit{"Chinese"} or \textit{"British"}, are often misclassified as locations despite referring to national identities.

\noindent \textbf{(4) Implied entities}. LLMs frequently misinterpret terms that imply specific entities. For example, product names like \textit{"iPhone"} or \textit{"Google Maps"} are often mislabeled as organizational entities due to their association with companies.

\noindent \textbf{(5) Entity omission and boundary errors.} LLMs struggle to recognize certain entities, such as abbreviations or long entities (e.g., long addresses). They may either omit these entities entirely or incorrectly segment them.

\section{Discussion}

The findings of our study highlight several potential directions for improving the performance of LLMs on financial NER tasks:

\textbf{Tuning LLMs for the Financial Domain.} A significant proportion of the observed failure cases involve domain-specific proper nouns. Fine-tuning LLMs with financial data could enhance their ability to accurately recognize such entities.

\textbf{Implementing self-correction strategies.} Our analysis in RQ2 identifies common mistakes made by LLMs in the FiNER-ORD task. Developing self-verification prompting strategies could allow LLMs to recognize and address these errors, thereby reducing recurrent failures.

\section{Conclusion}
This study presents the first systematic evaluations of generic LLMs in the FiNER-ORD task under different prompt designs, compared to state-of-the-art fine-tuned transformer-based models. Through comprehensive experiments with LLMs and their related lightweight versions, we demonstrate the capabilities and limitations of generic LLMs in handling domain-specific tasks. Our findings categorize five representative types of failures, along with their underlying causes. We release artifcats for future research \footnote{https://github.com/Alex-Lyu0419/Financial-Named-Entity-Recognition-How-Far-Can-LLM-Go}.

\bibliography{refs}

\begin{thebibliography}{19}
\providecommand{\natexlab}[1]{#1}

\bibitem[{Alvarado et~al.(2015)Alvarado, Verspoor, and Baldwin}]{alvarado2015domain}
Julio Cesar~Salinas Alvarado, Karin Verspoor, and Timothy Baldwin. 2015.
\newblock Domain adaption of named entity recognition to support credit risk assessment.
\newblock In \emph{Proceedings of the Australasian Language Technology Association Workshop 2015}, pages 84--90.

\bibitem[{Ashtiani and Raahemi(2023)}]{ashtiani2023news}
Matin~N Ashtiani and Bijan Raahemi. 2023.
\newblock News-based intelligent prediction of financial markets using text mining and machine learning: A systematic literature review.
\newblock \emph{Expert Systems with Applications}, 217:119509.

\bibitem[{Christiano et~al.(2017)Christiano, Leike, Brown, Martic, Legg, and Amodei}]{christiano2017deep}
Paul~F Christiano, Jan Leike, Tom Brown, Miljan Martic, Shane Legg, and Dario Amodei. 2017.
\newblock Deep reinforcement learning from human preferences.
\newblock \emph{Advances in neural information processing systems}, 30.

\bibitem[{{DeepInfra}(2024)}]{deepinfra}
{DeepInfra}. 2024.
\newblock {Deep Infra model cards}.
\newblock \url{https://deepinfra.com/models}.
\newblock Accessed: 2024-11-10.

\bibitem[{Devlin(2018)}]{devlin2018bert}
Jacob Devlin. 2018.
\newblock Bert: Pre-training of deep bidirectional transformers for language understanding.
\newblock \emph{arXiv preprint arXiv:1810.04805}.

\bibitem[{Dong et~al.(2022)Dong, Li, Dai, Zheng, Ma, Li, Xia, Xu, Wu, Liu et~al.}]{dong2022survey}
Qingxiu Dong, Lei Li, Damai Dai, Ce~Zheng, Jingyuan Ma, Rui Li, Heming Xia, Jingjing Xu, Zhiyong Wu, Tianyu Liu, et~al. 2022.
\newblock A survey on in-context learning.
\newblock \emph{arXiv preprint arXiv:2301.00234}.

\bibitem[{Dubey et~al.(2024)Dubey, Jauhri, Pandey, Kadian, Al-Dahle, Letman, Mathur, Schelten, Yang, Fan et~al.}]{dubey2024llama}
Abhimanyu Dubey, Abhinav Jauhri, Abhinav Pandey, Abhishek Kadian, Ahmad Al-Dahle, Aiesha Letman, Akhil Mathur, Alan Schelten, Amy Yang, Angela Fan, et~al. 2024.
\newblock The llama 3 herd of models.
\newblock \emph{arXiv preprint arXiv:2407.21783}.

\bibitem[{Google(2024)}]{gemini2024api}
Google. 2024.
\newblock {Gemini API}.
\newblock \url{https://ai.google.dev/gemini-api}.
\newblock Accessed: 2024-11-20.

\bibitem[{{Google}(2024)}]{googlecolab}
{Google}. 2024.
\newblock {Google Colaboratory}.
\newblock \url{https://colab.research.google.com/}.
\newblock Accessed: 2024-11-15.

\bibitem[{Li et~al.(2023{\natexlab{a}})Li, Chan, Zhu, Pei, Ma, Liu, and Shah}]{li2023chatgpt}
Xianzhi Li, Samuel Chan, Xiaodan Zhu, Yulong Pei, Zhiqiang Ma, Xiaomo Liu, and Sameena Shah. 2023{\natexlab{a}}.
\newblock {Are ChatGPT and GPT-4 general-purpose solvers for financial text analytics? A study on several typical tasks}.
\newblock \emph{arXiv preprint arXiv:2305.05862}.

\bibitem[{Li et~al.(2023{\natexlab{b}})Li, Wang, Ding, and Chen}]{li2023large}
Yinheng Li, Shaofei Wang, Han Ding, and Hang Chen. 2023{\natexlab{b}}.
\newblock Large language models in finance: A survey.
\newblock In \emph{Proceedings of the fourth ACM international conference on AI in finance}, pages 374--382.

\bibitem[{Liu(2019)}]{liu2019roberta}
Yinhan Liu. 2019.
\newblock Roberta: A robustly optimized bert pretraining approach.
\newblock \emph{arXiv preprint arXiv:1907.11692}, 364.

\bibitem[{OpenAI(2024)}]{openai2024chatgpt}
OpenAI. 2024.
\newblock {GPT-4o}.
\newblock \url{https://chat.openai.com}.
\newblock Accessed: 2024-10-24.

\bibitem[{{OpenAI}(2024)}]{openaichatgptguide}
{OpenAI}. 2024.
\newblock {Vision Guide}.
\newblock \url{https://platform.openai.com/docs/guides/vision}.
\newblock Accessed: 2024-10-24.

\bibitem[{Qin et~al.(2024)Qin, Chen, Feng, Wu, Zhang, Li, Li, Che, and Yu}]{qin2024large}
Libo Qin, Qiguang Chen, Xiachong Feng, Yang Wu, Yongheng Zhang, Yinghui Li, Min Li, Wanxiang Che, and Philip~S Yu. 2024.
\newblock Large language models meet nlp: A survey.
\newblock \emph{arXiv preprint arXiv:2405.12819}.

\bibitem[{Shah et~al.(2023)Shah, Vithani, Gullapalli, and Chava}]{shah2023finer}
Agam Shah, Ruchit Vithani, Abhinav Gullapalli, and Sudheer Chava. 2023.
\newblock Finer: Financial named entity recognition dataset and weak-supervision model.
\newblock \emph{arXiv preprint arXiv:2302.11157}.

\bibitem[{Wei et~al.(2022)Wei, Wang, Schuurmans, Bosma, Xia, Chi, Le, Zhou et~al.}]{wei2022chain}
Jason Wei, Xuezhi Wang, Dale Schuurmans, Maarten Bosma, Fei Xia, Ed~Chi, Quoc~V Le, Denny Zhou, et~al. 2022.
\newblock Chain-of-thought prompting elicits reasoning in large language models.
\newblock \emph{Advances in neural information processing systems}, 35:24824--24837.

\bibitem[{Wolf et~al.(2020)Wolf, Debut, Sanh, Chaumond, Delangue, Moi, Cistac, Rault, Louf, Funtowicz et~al.}]{wolf2020transformers}
Thomas Wolf, Lysandre Debut, Victor Sanh, Julien Chaumond, Clement Delangue, Anthony Moi, Pierric Cistac, Tim Rault, R{\'e}mi Louf, Morgan Funtowicz, et~al. 2020.
\newblock Transformers: State-of-the-art natural language processing.
\newblock In \emph{Proceedings of the 2020 conference on empirical methods in natural language processing: system demonstrations}, pages 38--45.

\bibitem[{Xie et~al.(2024)Xie, Han, Zhang, Lai, Peng, Lopez-Lira, and Huang}]{xie2024pixiu}
Qianqian Xie, Weiguang Han, Xiao Zhang, Yanzhao Lai, Min Peng, Alejandro Lopez-Lira, and Jimin Huang. 2024.
\newblock Pixiu: A comprehensive benchmark, instruction dataset and large language model for finance.
\newblock \emph{Advances in Neural Information Processing Systems}, 36.

\end{thebibliography}

\end{document}